# Building a Large-Scale Knowledge Base for Machine Translation

Kevin Knight and Steve K. Luk
USC/Information Sciences Institute
4676 Admiralty Way
Marina del Rey, CA 90292
{knight,luk}@isi.edu

## Abstract

Knowledge-based machine translation (KBMT) systems have achieved excellent results in constrained domains, but have not yet scaled up to newspaper text. The reason is that knowledge resources (lexicons, grammar rules, world models) must be painstakingly handcrafted from scratch. One of the hypotheses being tested in the PANGLOSS machine translation project is whether or not these resources can be semi-automatically acquired on a very large scale.

This paper focuses on the construction of a large ontology (or knowledge base, or world model) for supporting KBMT. It contains representations for some 70,000 commonly encountered objects, processes, qualities, and relations. The ontology was constructed by merging various online dictionaries, semantic networks, and bilingual resources, through semi-automatic methods. Some of these methods (e.g., conceptual matching of semantic taxonomies) are broadly applicable to problems of importing/exporting knowledge from one KB to another. Other methods (e.g., bilingual matching) allow a knowledge engineer to build up an index to a KB in a second language, such as Spanish or Japanese.

## Introduction

The PANGLOSS project is a three-site collaborative effort to build a large-scale knowledge-based machine translation system. Key components of PANGLOSS include New Mexico State University's Panglyzer parser (Farwell & Wilks 1991), Carnegie Mellon's translator's workstation (Frederking et al. 1993), and USC/ISI's PENMAN English generation system (Penman 1989). All of these systems combine to form a prototype Spanish-English translation system.

Another key component is the PANGLOSS ontology, a large-scale conceptual network for supporting semantic processing in other PANGLOSS modules. This network contains tens of thousands of nodes representing commonly encountered objects, entities, qualities, and relations. The upper (more abstract) region of the ontology is called the Ontology Base (OB) and consists of approximately 400 items that represent generalizations essential for the various PANGLOSS modules' linguistic processing during translation. The middle region of the ontology, approximately 50,000 items, provides a framework for a generic world model, containing items representing many English word senses. The lower (more specific) regions of the ontology provide anchor points for different application domains.

The purpose of the ontology is two-fold. First, it provides a common inventory of semantic tokens, used in both analysis and generation. These tokens form the bulk of the "lexicon" of the interlingua language. Second, the ontology describes which tokens are naturally related to which others, and in what ways, in our particular world. These relations form the "grammar" of the interlingua, where "grammaticality" of an interlingua sentence is identified with semantic plausibility.

Because large-scale knowledge bases are difficult to build by hand, we have chosen to pursue primarily semi-automatic methods for manipulating and merging existing resources. The next section sketches out the information in five such resources, and subsequent sections describe algorithms for extracting and merging this information.

## Linguistic Resources

We selected the following resources with the idea that each contains a piece of the puzzle we are trying to build: (1) the PENMAN Upper Model from USC/ISI, (2) the ONTOS model from Carnegie Mellon University, (3) the Longman's Dictionary of Contemporary English (LDOCE), (4) WordNet, and (5) the Harper-Collins Spanish-English bilingual dictionary.

### PENMAN Upper Model

The Upper Model (Bateman 1990) is a top-level network of about 200 nodes, implemented in the LOOM knowledge representation language (MacGregor 1988), and used by the PENMAN English generation system (Penman 1989) to drive its linguistic choices. PENMAN makes extensive use of syntactic-semantic correspondences; if a concept is taxonomized under a particular node of the Upper Model, then an English word referring to that concept will have a particular set of default grammatical behaviors. Exceptions are coded in the lexicon.

## ONTOS

Of comparable size to the PENMAN Upper Model, ONTOS (Carlson & Nirenburg 1990) is a top-level ontology designed to support machine translation. The event structure is based on cross-linguistic studies of verbs, and case roles and filler restrictions are represented independently of any particular language. ONTOS also includes object hierarchies, scalar attributes, and complex events.

## Longman's Dictionary (LDOCE)

LDOCE is a learner's dictionary of English with 27,758 words and 74,113 word senses. Each word sense comes with:

- A short definition. One of the unique features of LDOCE is that its definitions only use words from a "control vocabulary" list of 2000 words. This makes it attractive from the point of view of extracting semantic information by parsing dictionary entries.
- Examples of usage.
- One or more of 81 syntactic codes (e.g., [B3]: adj followed by *to*).
- For nouns, one of 33 semantic codes (e.g., [H]: human).
- For nouns, one of 124 pragmatic codes (e.g., [ECZB]: economics/business).

Another important feature of LDOCE is that its sense identifiers are used in the semantic fields of a medium-scale Spanish lexicon built by hand at New Mexico State University as part of PANGLOSS.

## WordNet

WordNet (Miller 1990) is a semantic word database based on psycholinguistic principles. It is a large-scale resource like LDOCE, but its information is organized in a completely different manner. WordNet groups synonymous word senses into single units ("synsets"). Noun senses are organized into a deep hierarchy, and the database also contains part-of links, antonym links, and others. Approximately half of WordNet synsets have brief informal definitions.

## Collins Bilingual Dictionary

The Harper-Collins Bilingual Spanish-English dictionary (Collins 1971) contains tens of thousands of Spanish headwords and English translations. Like words in LDOCE definitions, word translations are not marked by sense, but they are sometimes annotated with subject field codes, such as Military [MIL] or Commercial [COM].

## Merging Resources

Our initial goal was to combine all of these resources into a conceptual network of about 50,000 nodes, indexed by structured lexicons for both English and Spanish. This network drives the PENMAN generator and, to the extent that it can, helps in semantic disambiguation tasks during parsing.[1]

Figure 1 shows the plan of attack. The PENMAN Upper Model and ONTOS were merged by hand to create the Ontology Base (OB). This structure continues to undergo revision as we add case roles and other support for the interlingua. WordNet was then subordinated/merged into the OB. The result is a large knowledge base in which most concepts are named by WordNet names, but in which some have three names, one each from Ontos, the Upper Model, and WordNet. Proper taxonomization under the Ontology Base ensures the proper working of PENMAN, since the PENMAN Upper Model is embedded there intact. The subordination of WordNet involved breaking the network into some 200 pieces and merging each manually into the OB.

The next step was to merge word senses from LDOCE with those of WordNet. There were several motivations for doing this: (1) LDOCE has a great deal of lexical information missing from WordNet, including syntactic and subject field codes, and controlled-vocabulary definitions; and (2) LDOCE sense identifiers are legal tokens in the PANGLOSS interlingua, as much of the ULTRA Spanish lexicon is written in terms of these identifiers. Merging LDOCE and WordNet senses is a very large task, for which we developed semi-automatic algorithms.

The final step was to build up a large Spanish lexicon for the ontology. Again, doing this manually was too expensive, so we built algorithms for extracting a lexicon from the Collins bilingual dictionary semi-automatically.

Each resource makes its own contributions to the final product. LDOCE offers syntax and subject area, WordNet offers synonyms and hierarchical structuring, the upper structures organize the knowledge for natural language processing in general and English generation in particular, and finally, the bilingual dictionary lets us index the ontology from a second language. The bulk of the rest of this paper is devoted to the three automatic merging algorithms developed in support of the work in Figure 1. The first two algorithms support the LDOCE-WordNet merge, while the third supports the Collins-Ontology merge.

## Definition Match Algorithm

The Definition Match algorithm is based on the idea that two word senses should be matched if their two

---

[1] Disambiguation algorithms are described in a separate paper (Luk 1994).

definitions share words. For example, there are two noun definitions of "batter" in LDOCE:

- (batter_2_0) "mixture of flour, eggs, and milk, beaten together and used in cooking"
- (batter_3_0) "a person who bats, esp in baseball — compare BATSMAN"

and two definitions in WordNet:

- (BATTER-1) "ballplayer who bats"
- (BATTER-2) "a flour mixture thin enough to pour or drop from a spoon"

The Definition Match algorithm will match (batter_2_0) with (BATTER-2) because their definitions share words like "flour" and "mixture." Similarly (batter_3_0) and (BATTER-1) both contain the word "bats," so they are also matched together.

Not all senses in WordNet have definitions, but most have synonyms and superordinates. For this reason, the algorithm looks not only at WordNet definitions, but also at locally related words and senses. For example, if synonyms of WordNet sense $x$ appear in the definition of LDOCE sense $y$, then this is evidence that $x$ and $y$ should be matched.

The complete Definition Match algorithm is given in (Knight 1993). Here we give a brief sketch. Given a word $w$, we identify and stem all open-class content words from definitions and example sentences of $w$ in both dictionaries. We add to this set all synonyms, superordinates, siblings, and super-superordinates from all senses of $w$ in WordNet. The set is then reduced to contain only words that cropped up in both resources, minus $w$ itself. The next step is to create a two-dimensional matrix for each resource. For LDOCE, $L[i, x]$ is set to 1.0 just in case word $x$ appears in the definition of sense $i$ (otherwise, it is set to 0.01). For WordNet, $W[x, j]$ is set to 1.0 if $x$ is a synonym or superordinate of sense $j$, 0.8 if $x$ is in the definition of sense $j$, 0.6 if $x$ is a sibling or grandparent of sense $j$, and 0.01 otherwise. Multiplying matrices L and W yields a similarity matrix SIM. We repeatedly choose the largest value $v$ in the SIM matrix, using the indices $i$ and $j$ of that value to propose a match between LDOCE sense $i$ and WordNet sense $j$ of word $w$ (at confidence level $v$).

Empirical results are as follows. We ran the algorithm over all nouns in both LDOCE and WordNet. We judged the correctness of its proposed matches, keeping records of the confidence levels and the degree of ambiguity present. For low-ambiguity words (with exactly two senses in LDOCE and two in WordNet), the results are:

| confidence level | pct. correct | pct. coverage |
| --- | --- | --- |
| $\geq 0.0$ | 75% | 100% |
| $\geq 0.4$ | 85% | 53% |
| $\geq 0.8$ | 90% | 27% |

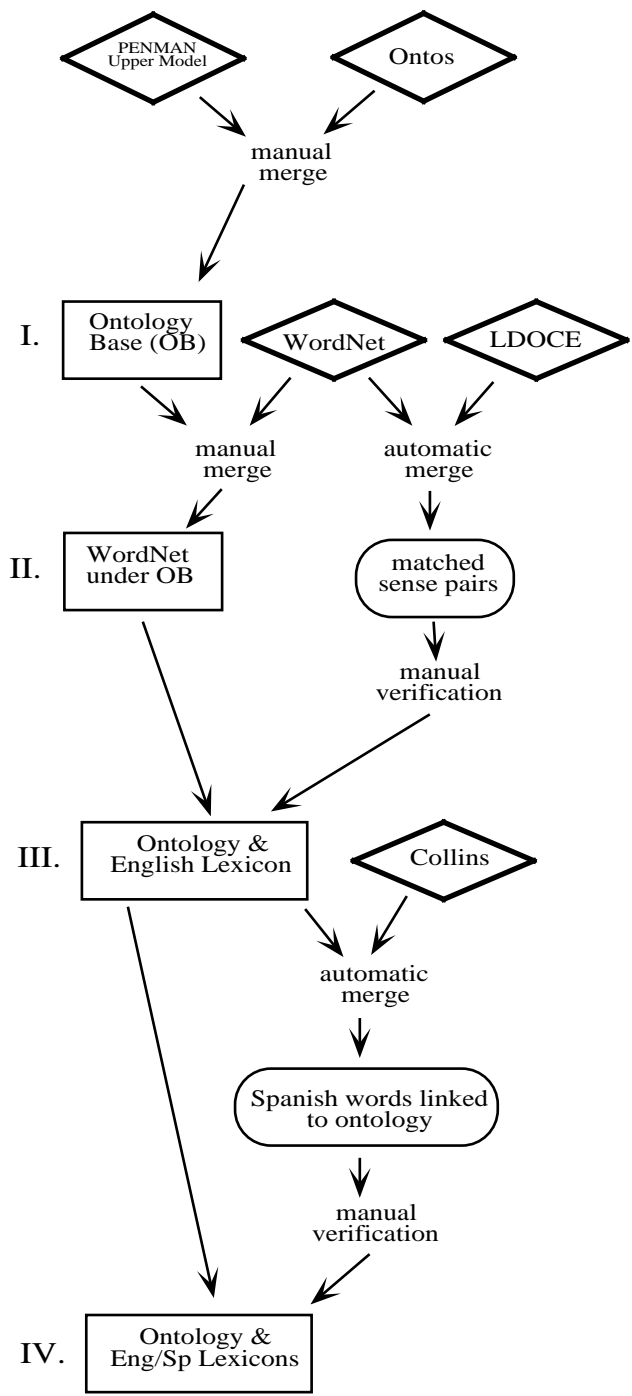

Figure 1: Merging Information in Five Linguistic Resources to Build a Large Scale Ontology for Machine Translation

At confidence levels ≥ 0.0, 75% of the proposed matches are correct. If we restrict ourselves to only matches proposed at confidence ≥ 0.8, accuracy increases to 90%, but we only get 27% of the possible matches.

For high-ambiguity words (more than five senses in LDOCE and WordNet), the results are:

| confidence level | pct. correct | pct. coverage |
|---|---|---|
| ≥ 0.0 | 47% | 100% |
| ≥ 0.1 | 76% | 44% |
| ≥ 0.2 | 81% | 20% |

Accuracy here is worse, but increases sharply when we only consider high confidence matches.

The algorithm's performance is reasonable, given that 45% of WordNet senses have no definitions and that many existing definitions are brief and contain misspellings. Still, there are several improvements to be made—e.g., modify the "greedy" strategy in which matches are extracted from SIM matrix, weigh rare words in definitions more highly than common ones, and/or score senses with long definitions lower than ones with short definitions. These improvements yield only slightly better results, however, because most failures are simply due to the fact that matching sense definitions often have no words in common.

## Hierarchy Match Algorithm

The Hierarchy Match algorithm dispenses with sense definitions altogether. Instead, it uses the various sense hierarchies inside LDOCE and WordNet.

WordNet noun senses are arranged in a deep is-a hierarchy. For example, SEAL-7 is a PINNIPED-1, which is on AQUATIC-MAMMAL-1, which is a EUTHERIAN-1, which is a MAMMAL-1, which is ultimately an ANIMAL-1, and so forth.

LDOCE has two fairly flat hierarchies. The *semantic code* hierarchy is induced by a set of 33 semantic codes drawn up by Longman lexicographers. Each sense is marked with one of these codes, e.g., "H" for human "P" for plant, "J" for movable object. The other hierarchy is the *genus sense* hierarchy. (Bruce & Guthrie 1992) have built an automatic algorithm for locating and disambiguating genus terms (head nouns) in sense definitions. For example, (bat_1_1) is defined as "any of the several types of specially shaped wooden stick used for …" The genus term for (bat_1_1) is (stick_1_1). The genus sense and the semantic code hierarchies were extracted automatically from LDOCE. The semantic code hierarchy is fairly robust, but since the genus sense hierarchy was generated heuristically, it is only 80% correct.

The idea of the Hierarchy Match algorithm is that once two senses are matched, it is a good idea to look at their respective ancestors and descendants for further matches. For example, once (animal_1_2) and ANIMAL-1 are matched, we can look into their respective animal-subhierarchies. We find that the word "seal" is locally unambiguous—only one sense of "seal" refers to an animal (in both LDOCE and WordNet). So we feel confident to match those seal-animal senses. As another example, suppose we know that (swan_dive_0_0) is the same concept as (SWAN-DIVE-1). We can then match their superordinates (dive_2_1) and (DIVE-3) with high confidence; we need not consider other senses of "dive."

Here is the algorithm:

1. Initialize the set of matches:

 (a) Retrieve all words that are unambiguous in both LDOCE and WordNet. Match their corresponding senses, and place all the matches on a list called M1.

 (b) Retrieve a prepared list of hand-crafted matches. Place these matches on a list called M2. We created 15 of these, mostly high-level matches like (person_0_1, PERSON-2) and (plant_2_1, PLANT-3). This step is not strictly necessary, but provides guidance to the algorithm.

2. Repeat until M1 and M2 are empty:

 (a) For each match on M2, look for words that are unambiguous within the hierarchies rooted at the two matched senses. Match the senses of locally unambiguous words and place the matches on M1.

 (b) Move all matches from M2 to a list called M3.

 (c) For each match on M1, look upward in the two hierarchies from the matched senses. Whenever a word appears in both hierarchies, match the corresponding senses, and place the match on M2.

 (d) Move all matches from M1 to M2.

The algorithm operate in phases, shifting matches from M1 to M2 to M3, placing newly-generated matches on M1 and M2. Once M1 and M2 are exhausted, M3 contains the final list of matches proposed by the algorithm. Again, we can measure the success of the algorithm along two dimensions, coverage and correctness:

| phase | pct. correct | matches proposed |
|---|---|---|
| Step 1 | 99% | 7563 |
| Step 2(a) | 94% | 876 |
| Step 2(c) | 85% | 530 |
| Step 2(a) | 93% | 2018 |
| Step 2(c) | 83% | 40 |
| Step 2(a) | 92% | 99 |
| Step 2(c) | 100% | 2 |

In the end, the algorithm produced 11,128 noun sense matches at 96% accuracy. We expected 100% accuracy, but the algorithm was foiled at several places by errors in one or another of the hierarchies. For example, (savings_bank_0_0) is mistakenly a subclass of river bank (bank_1_1) in the LDOCE genus hierarchy, rather than (bank_1_4), the financial institution. "Savings bank" senses are matched in step 1(a), so step 2(c) erroneously goes on to match the river bank of LDOCE with the financial institution of WordNet.

Fortunately, the Definition and Hierarchy Match algorithms complement one another, and there are several ways to combine them. Our practical experience has been to run the Hierarchy Match algorithm to completion, remove the matched senses from the databases, then run the Definition Match algorithm.

## Bilingual Match Algorithm

The goal of this algorithm is annotate our ontology with a large Spanish lexicon. This lexicon will of course be fairly rough, with some senses not matching up exactly, and with no lexical decomposition. But getting a large-scale lexicon up and running shows us where the real MT problems are; we don't have to imagine them.

The raw materials we have to work with are: (1) mappings between Spanish and English words, from Collins bilingual dictionary, (2) mappings between English words and ontological entities, primarily from WordNet, and (3) conceptual relations between ontological entities. What we do not yet have are direct links between Spanish words and ontological entities. Consider that the Spanish word *manzana* can be translated as *block* in English; however, *manzana* only maps to one of the concepts referred to by *block*, namely CITY-BLOCK. It does not map to BUILDING-BLOCK. So our task is one of disambiguating each English word in the list of possible translations.

Fortunately, the bilingual dictionary provides a bit more structure to exploit. The entry for *banco* looks roughly like:

```
banco. nm.  bench, seat;
            bank, shoal;
            school, shoal;
            layer, stratum;
            bank [COM];
               ...
```

We can use the division into senses (by semicolons), the synonyms given for each sense, and the subject field codes annotating some senses.

We take advantage of synonyms by using WordNet's synsets and hierarchies. The words *school* and *shoal* each have many meanings in WordNet, but only one pair of meanings coincide at the same WordNet synset. So we are able to perform disambiguation and map *banco* onto SCHOOL-OF-FISH rather than SCHOOL-FOR-KIDS. The words *bench* and *seat* are not synonyms in WordNet, but there is a pair of senses that are very close to each other; they share a common immediate parent in the hierarchy. The Bilingual Match algorithm postulates mappings between *banco* and BENCH-FOR-SITTING and SEAT-FOR-SITTING, but at a slightly lower level of confidence than the one for SCHOOL-OF-FISH. We penalize the proposed match a constant factor for each link traversed in WordNet to reach a common parent node.[2]

Sometimes only one English word is given as a translation. If the word is unambiguous, we postulate the match at high confidence. Otherwise, we try to make use of Collins subject field codes, as in *banco = bank* [COM]. Fortunately, because we have merged LDOCE and WordNet, our ontology concepts are annotated with LDOCE subject field codes that are similar to the ones found in Collins. Rather than compile a correspondence table of Collins-LDOCE field codes by hand, we generated such a table automatically. For each word mapping from Spanish to English, we considered every meaning of the English word, entering all field code matches into the table. Due to ambiguity, some spurious matches were entered, such as [ZOOL] for *palo*, which in English means *bat* as in baseball. (ZOOLogy was picked up from the flying-mammal sense of *bat*). However, spurious matches were largely eliminated when we removed field code matches that occurred less than six times. Once we built the table, we put it to use in disambiguating English word translations in Collins. If a word is marked with a Collins field code, we simply look for ontology items marked with corresponding LDOCE field codes. Accuracy figures for the Bilingual Match algorithm are now being computed, as human verifiers proceed through the 50,000 proposed mappings from Spanish words to the ontology.

## Discussion

For each of the merge algorithms described above, we have built a simple interface that allows a person to verify and/or correct the results generated. The verification interface places the proposed match at the top of a list of alternatives. If the proposed match is correct, the verifier need look no further; the set up is much like a spelling correction interface that sorts alternatives by likelihood rather than, say, alphabetic order. The principle here is that humans are much faster at verifying information than generating it from scratch.

Semi-automatic merging brings together complementary sources of information. It also allows us to detect errors and omissions where the resources are redundant. For example, after the WordNet-LDOCE merge was verified, we were able to automatically locate hundreds of inconsistencies between the WordNet and LDOCE (genus-sense) hierarchies. Many inconsistencies pointed to errors in the genus-word identification or genus-sense disambiguation, while others pointed to different taxonomic organizations that can be merged into one lattice structure. Another benefit of merging resources is that it makes subsequent knowl-edge acquisition easier. For example, in designing the Bilingual Match algorithm, we were free to make use of information in both WordNet and LDOCE.

---

[2] A more elaborate scheme would weight links, as in (Resnik 1993).

## Related Work and Future Work

Automatic dictionary merging is an old line of research; recent work includes (Klavans & Tzoukermann 1990; Klavans 1990; Rohini & Burhans 1994). Many times, the dictionaries merged were roughly similar, while in our work, we have chosen three very different resources. Another motivation for merging dictionaries is to get several definitions for the same sense, to maximize the information that can be extracted by analyzing those definitions. We have not yet extracted information from LDOCE definitions, though this is a clear source of knowledge for enriching the ontology, and there is a great deal of fine work to build on (Klavans *et al.* 1991; Wilks *et al.* 1990; Klavans, Chodorow, & Wacholder 1992). Our other source of knowledge is free text, and we are currently exploring techniques for automatically extracting semantic constraints (Luk 1994). (Okumura & Hovy 1994) use ideas related to the bilingual match algorithm to semi-automatically construct a Japanese lexicon for the PANGLOSS ontology.

## Acknowledgments

We would like to thank Eduard Hovy for his support and for comments on a draft of this paper. Yolanda Gil also provided useful comments. Thanks to Richard Whitney for significant assistance in programming and verification. The Ontology Base was built by Eduard Hovy, Licheng Zeng, Akitoshi Okumura, Richard Whitney, and Kevin Knight. Gratitude goes to Longman Group, Ltd., for making the machine readable version of LDOCE available to us, and to HarperCollins Publishers for letting us experiment with their bilingual dictionary. Louise Guthrie assisted in LDOCE/Collins extraction and kindly provided us with the LDOCE genus sense hierarchy. This work was carried out under ARPA Order No. 8073, contract MDA904-91-C-5224.